\documentclass[12pt,a4paper,onecolumn,twoside,openright]{ranlp_bk}

\usepackage{times}

\usepackage{epsf}        
\usepackage{epsfig}

\newcommand{\ranlptitle}[1]{    \chapter*{#1}
                                \setcounter{section}{0}
                                \setcounter{figure}{0}
                                \setcounter{table}{0}
                                \setcounter{equation}{0}
                                \setcounter{footnote}{0}   }

\newcommand{\ranlpcontents}[2]{\addcontentsline{toc}{chapter}{\emph{#1}\\#2}}

\renewcommand{\author}[1]{\vspace*{-2mm}
                          \begin{center}\textsc{#1}\end{center}}

\newcommand{\affiliation}[1]{\begin{center}\emph{#1}\end{center}
                             \vspace*{3mm}}

\newcommand{\ranlpaddr}[1]{}    

\newcommand{\runningheads}[2]{  
        \markboth{\small\uppercase{#1}}{\small\uppercase{#2}}}

\newcommand{\ranlpabstract}[1]{   \centerline{\textbf{Abstract}}\vspace*{-1mm}
                                  \begin{quote}\small #1 \end{quote} }

\renewcommand{\figurename}{Fig.} 

\renewenvironment{thebibliography}{
        \vspace{1.7em}\small \centerline{\bf REFERENCES}\vspace*{1mm}
        \begin{list}{}
                {\setlength{\itemsep}{-0.6mm}
                 \setlength{\itemindent}{-7mm}
                 \setlength{\leftmargin}{7mm} 
                }               }{\end{list}}

\makeatletter
\long\def\@makefntext#1{\@setpar{\@@par\@tempdima \hsize 
  \advance\@tempdima-10pt\parshape \@ne 10pt \@tempdima}\par
  \parindent 1em\noindent \hbox to \z@{\hss$^{\@thefnmark}\ $}#1}
\makeatother

\renewcommand{\indexname}{Index of Subjects and Terms}

\newcommand{\hush}[1]{}             

\newlength{\spacing}
\newcommand{\nspace}[1]{\setlength{\baselineskip}{#1\spacing}}

\begin{document}
   \pagenumbering{arabic}


\ranlptitle{Combining Independent Modules \\
            in Lexical Multiple-Choice Problems}

\author{Peter D. Turney$^*$, Michael L. Littman$^{**}$, \\
        Jeffrey Bigham$^\star$, \& Victor Shnayder$^{\star\star}$}

\affiliation{$^*$NRC, 
             $^{**}$Rutgers Univ.,
             $^\star$Univ. of Washington,
             $^{\star\star}$Harvard Univ.}

\ranlpcontents{Peter D. Turney, Michael L. Littman, Jeffrey Bigham \& \\ Victor Shnayder}
              {Combining independent modules in lexical multiple-choice \\ problems}

\ranlpaddr{
Peter D. Turney \\
Inst. of Information Tech. \\
National Research Council of Canada \\
Ont., Canada, K1A 0R6 \\
\texttt{peter.turney@nrc.ca} \\
}
\ranlpaddr{
Michael L. Littman \\
Dept. of Computer Science \\
Rutgers University \\
Piscataway, NJ 08854-8019 \\
\texttt{mlittman@cs.rutgers.edu} \\
}
\ranlpaddr{
Jeffrey Bigham \\
Computer Science \& Engineering \\
University of Washington \\
Seattle, WA 98195-2350 \\
\texttt{jbigham@u.washington.edu} \\
}
\ranlpaddr{
Victor Shnayder \\
Division of Engineering and Applied Sciences \\
Harvard University \\
Cambridge, MA 02138 \\
\texttt{shnayder@eecs.harvard.edu} \\
}

\runningheads{TURNEY, \ LITTMAN, \ BIGHAM, \ \& \ SHNAYDER}
             {COMBINING \ INDEPENDENT \ MODULES}
             
\ranlpabstract{Existing statistical approaches to natural language problems are very
coarse approximations to the true complexity of language processing.
As such, no single technique will be best for all problem
instances.  Many researchers are examining ensemble methods that
combine the output of multiple modules to
create more accurate solutions.  This paper examines three merging
rules for combining probability distributions: the familiar mixture
rule, the logarithmic rule, and a novel product rule.
These rules were
applied with state-of-the-art results to two
problems used to assess human mastery of lexical
semantics --- synonym questions and analogy questions.  All three
merging rules result in ensembles that are more accurate than any of
their component modules.  The differences among the three rules are not statistically
significant, but it is suggestive that the popular mixture rule
is not the best rule for either of the two problems.}

\section{Introduction}

Asked to articulate the relationship between the words {\small\sf broad}
and {\small\sf road}, you might consider a number of 
possibilities.  Orthographically,
the second can be
derived from the first by deleting the initial letter,
while semantically, the first can modify the second to
indicate above-average width.
Many possible relationships would need to be considered, depending on
the context.  In addition, many different computational approaches could
be brought to bear, leaving a designer of a natural language
processing system with some difficult choices.  A sound software
engineering approach is to develop separate modules using independent
strategies, then to combine the output of the modules to produce a
unified solver.

The concrete problem we consider here is predicting the correct answers to
multiple-choice questions. \index{multiple-choice questions} 
Each instance consists of a context and a
finite set of choices, one of which is correct.  Modules produce a
probability distribution over the choices and a merging rule is used
to combine these distributions into one.
This distribution, along with relevant utilities, can then be used to
select a 
candidate answer from the set of choices.  The merging rules we
considered are parameterized, and we set parameters by a maximum
likelihood approach on a collection of training instances.

Many problems can be cast in a multiple-choice framework, including
optical digit recognition
(choices are the 10 digits), 
word
sense disambiguation
(choices are a word's possible senses), 
text categorization (choices are the classes),
and part-of-speech tagging
(choices are the grammatical categories).
This paper looks at multiple-choice
synonym
questions (part of the Test of English as a Foreign Language) 
\index{Test of English as a Foreign Language (\textsc{toefl})}
and multiple-choice verbal analogy questions (part of the 
SAT college entrance exam).\index{sat@\textsc{sat}}
Recent work has shown that algorithms for 
solving multiple-choice synonym questions can be used to determine the
\emph{semantic orientation} of a word; i.e., whether the word
conveys praise or criticism~(Turney \& Littman 2003b).  Other research
establishes that algorithms for solving multiple-choice verbal analogy
questions can be used to ascertain the \emph{semantic relation}
in a noun-modifier expression; e.g., in the expression
``laser printer'', the modifier ``laser'' is an \emph{instrument}
used by the noun ``printer'' (Turney \& Littman 2003a).

The paper offers two main contributions.  First, it introduces and
evaluates several new modules for answering multiple-choice synonym
questions and verbal analogy questions.  Second, it presents a novel
product rule for combining module outputs and compares it with other
similar merging rules.

Section~\ref{turney-s:problem} formalizes the problem addressed in this paper
and introduces the three merging rules we study in detail: the mixture
rule, the logarithmic rule, and the product rule.
Section~\ref{turney-s:synonyms} presents empirical results on
synonym problems and Section~\ref{turney-s:analogies} considers analogy problems.
Section~\ref{turney-s:conclusion} summarizes and wraps up.  

\section{Module combination}
\label{turney-s:problem}
\index{module combination}

The following synonym question is a typical multiple-choice question:
{\small\sf hidden}:: (a) {\small\sf laughable}, (b) {\small\sf veiled}, (c)
{\small\sf ancient}, (d) {\small\sf revealed}.  The stem, {\small\sf hidden}, is the
question.  There are $k=4$ choices, and the question writer asserts
that exactly one (in this case, (b)) has the same meaning as the stem word.
The accuracy of a solver is measured by its fraction of
correct answers on a set of $\ell$ testing instances.

In our setup, knowledge about the multiple-choice task is encapsulated
in a set of $n$ modules, each of which can take a question instance
and return a probability distribution over the $k$ choices.  For a
synonym task, one module might be a statistical approach that makes
judgments based on analyses of word co-occurrence, while another might
use a thesaurus to identify promising candidates.  These modules are
applied to a training set of $m$ instances, producing probabilistic
``forecasts''; $p^h_{ij} \ge 0$ represents the probability assigned by
module $1\le i \le n$ to choice $1 \le j \le k$ on training instance
$1 \le h \le m$.  The estimated probabilities are distributions of the
choices for each module $i$ on each instance $h$: $\sum_j p^h_{ij} =
1$.

\subsection{Merging rules} 
\index{merging rules}

The rules we considered are parameterized by a set of weights
$w_i$, one for each module.  For a given merging rule, a setting of
the weight vector $w$ induces a probability distribution over the
choices for any instance.  Let $D^{h,w}_j$ be the probability assigned
by the merging rule to choice $j$ of training instance $h$ when the
weights are set to $w$.  Let $1 \le a(h) \le k$ be the correct answer
for instance $h$.  We set weights to maximize the likelihood of the
training data: $w = {\rm argmax}_{w'} \prod_h D^{h,w'}_{a(h)}$.  The same
weights maximize the \emph{mean likelihood}, the geometric mean of the
probabilities assigned to correct answers.

We focus on three merging rules in this paper.  The \emph{mixture
rule} \index{mixture rule}
combines module outputs using a weighted sum
and can be written $M^{h,w}_j = \sum_i w_i p^h_{ij},$
where $D^{h,w}_j = M^{h,w}_j / \sum_j M^{h,w}_j$ is the probability
assigned to choice $j$ of instance $h$ and $0\le
w_i\le 1$.  The rule can be justified by assuming each instance's answer
is generated by a single module chosen via the distribution
$w_i/\sum_i w_i$.

The \emph{logarithmic rule} \index{logarithmic rule}
combines the logarithm of module outputs
by $L^{h,w}_j = \exp(\sum_i w_i \ln p^h_{ij}) = \prod_i
(p^h_{ij})^{w_i}$, where $D^{h,w}_j = L^{h,w}_j / \sum_j L^{h,w}_j$ is
the probability the rule assigns to choice $j$ of instance $h$.  The
weight $w_i$ indicates how to scale the module probabilities before
they are combined multiplicatively.  Note that modules that output
zero probabilities must be modified before this rule can be used.

The \emph{product rule} \index{product rule}
can be written in the form
$P^{h,w}_j = \prod_i (w_i p^h_{ij} + (1-w_i)/k),$
where $D^{h,w}_j = P^{h,w}_j / \sum_j P^{h,w}_j$ is the probability
the rule assigns to choice $j$.  The weight $0\le
w_i \le 1$ indicates how module $i$'s output should be mixed with a uniform distribution
(or a prior, more generally) before outputs are combined
multiplicatively.  As with the mixture and logarithmic 
rules, a module with a weight of zero has no influence on the final
assignment of probabilities.  

For the experiments reported here, we adopted a straightforward
approach to finding the weight vector $w$ that maximizes the
likelihood of the data.  The weight optimizer reads in the output of
the modules, \index{weight optimization}
chooses a random starting point for the weights, then hillclimbs using
an approximation of the partial derivative.
Although more sophisticated optimization algorithms are well known, we
found that the simple discrete gradient approach worked well for our
application.

\subsection{Related work} 

Merging rules of various sorts have been studied for many years, and
have gained prominence recently for natural language applications.
Use of the mixture rule and its variations is quite common.  Recent
examples include the work of Brill \& Wu (1998) on part-of-speech
tagging, 
Littman et al. (2002) on crossword-puzzle clues and
Florian \& Yarowsky (2002) on a word-sense disambiguation task.  
We use the name ``mixture rule''
by analogy to the mixture of experts model~(Jacobs et al. 1991),
which combined expert opinions in an analogous way.  In the
forecasting literature, this rule is also known as the linear opinion
pool; Jacobs (1995) provides a summary of the theory and
applications of the mixture rule in this setting.

The logarithmic opinion pool of Heskes (1998) is the
basis for our logarithmic rule.  Boosting~(Schapire 1999) also 
uses a logistic-regression-like rule to
combine outputs of simple modules to perform state-of-the-art
classification.
The product of experts approach also combines distributions
multiplicatively, and Hinton (1999) argues that this is an
improvement over the ``vaguer'' probability judgments commonly
resulting from the mixture rule.  A survey by Xu et al. (1992) includes 
the equal-weights version of the mixture rule. A derivation of the
unweighted product rule appears in Xu et al. (1992) and 
Turney et al. (2003).  

An important contribution of the current work is the product rule,
which shares the simplicity of the mixture rule and the probabilistic
justification of the logarithmic rule.  We have not seen an analog of
this rule in the forecasting or learning literatures.

\section{Synonyms}
\label{turney-s:synonyms}
\index{synonyms}

We constructed a training set of 431 4-choice synonym
questions and randomly divided them into 331 training questions and 100
testing questions.
We created four modules, described next, and ran each module on the
training set.  We used the results to set the weights for the three
merging rules and evaluated the resulting synonym solver on
the test set. Module outputs, where applicable, were normalized to form a
probability distribution by scaling them to add to one before merging.

\subsection{Modules}

\textbf{LSA.} Following Landauer \& Dumais (1997), we used latent semantic
analysis to recognize synonyms.  Our LSA module queried the web
interface developed at the University of Colorado 
({\tt \small lsa.colorado.edu}), which has a
300-dimensional vector representation for each of tens of thousands of
words.

\textbf{PMI-IR.} Our Pointwise Mutual Information-Information
Retrieval module used the AltaVista search engine ({\tt \small www.altavista.com})
to determine the number of
web pages that contain the choice and stem in close proximity.  PMI-IR
used the third scoring method (near each other, but not near
{\small\sf not}) designed by Turney (2001), since it performed best
in this earlier study.

\textbf{Thesaurus.} Our Thesaurus module also used the web to measure
word pair similarity.  The module queried Wordsmyth 
({\tt \small www.wordsmyth.net}) and collected any words listed in
the ``Similar Words'', ``Synonyms'', ``Crossref. Syn.'', and ``Related
Words'' fields.  
The module created synonym lists for the stem and
for each choice, then scored them by their overlap.

\textbf{Connector.}
Our Connector module used summary pages from querying Google ({\tt \small
google.com}) with pairs of words to estimate pair similarity.
It assigned a score to a pair of words by
taking a weighted sum of both the number of times they
appear separated by one of the symbols {\bf [}, {\bf "}, {\bf :}, {\bf
,}, {\bf =}, {\bf /}, {\bf $\backslash$}, {\bf (}, {\bf ]},
{\small\sf means}, {\small\sf defined}, {\small\sf equals}, {\small\sf synonym},
whitespace, and {\small\sf and} and the number of times {\small\sf dictionary}
or {\small\sf thesaurus} appear anywhere in the Google summaries.

\begin{table}
\begin{small}
\begin{center}
\begin{tabular}{lcc}
Synonym solvers & Accuracy & Mean likelihood \\
 \hline
LSA only        & 43.8\%   & .2669 \\
PMI-IR only     & 69.0\%   & .2561 \\
Thesaurus only  & 69.6\%   & .5399 \\
Connector only  & 64.2\%   & .3757 \\
 \hline
All: mixture    & 80.2\%   & .5439 \\ 
All: logarithmic& 82.0\%   & .5977 \\ 
All: product    & 80.0\%   & .5889 \\ 
\end{tabular}
\end{center}
\end{small}
\caption{Comparison of results for merging rules on synonym
problems} 
\label{turney-t:synprob}
\vspace*{-1em}
\end{table}

\subsection{Results} 

Table~\ref{turney-t:synprob} presents the result of
training and testing each of the four modules on synonym problems.
The first four lines list the accuracy and mean likelihood obtained
using each module individually (using the product rule to set the
individual weight).  The highest accuracy is that of the
Thesaurus module at 69.6\%.  All three merging rules were able to
leverage the combination of the modules to improve performance to
roughly 80\% --- statistically significantly better than the best
individual module.

Although the accuracies of the merging rules are nearly
identical, the product and logarithmic rules assign higher
probabilities to correct answers, as evidenced by the mean likelihood.
To illustrate the decision-theoretic implications of this difference,
imagine using the probability judgments in a system that receives
a score of $+1$ for each right answer and $-1/2$ for each wrong
answer, but can skip questions. In this case, the system should make a guess whenever the
highest probability choice is above $1/3$.  For the test questions,
this translates to scores of 71.0 and 73.0 for the product and
logarithmic rules, but only 57.5 for the mixture rule; it skips many
more questions because it is insufficiently certain.

\begin{table}
\begin{small}
\begin{center}
\begin{tabular}{lcc}
Reference		& Accuracy & 95\% confidence \\
   \hline
Landauer \& Dumais (1997)    & 64.40\%  & 52.90--74.80\% \\
non-native speakers	         & 64.50\%  & 53.01--74.88\% \\
Turney (2001)		             & 73.75\%  & 62.71--82.96\% \\
Jarmasz \& Szpakowicz (2003) & 78.75\%  & 68.17--87.11\% \\
Terra \& Clarke (2003)       & 81.25\%  & 70.97--89.11\% \\
Product rule		             & 97.50\%  & 91.26--99.70\%
\end{tabular}
\end{center}
\end{small}
\caption{Published TOEFL synonym results} 
\label{turney-t:toefl}
\vspace*{-1em}
\end{table}

\subsection{Related work and discussion} 

Landauer \& Dumais (1997) introduced
the Test of English as a Foreign Language (TOEFL) 
\index{Test of English as a Foreign Language (\textsc{toefl})}
synonym task as a
way of assessing the accuracy of a learned representation of lexical
semantics.  Several studies have since used the same data set for
direct comparability; Table~\ref{turney-t:toefl} presents these results.

The accuracy of LSA~(Landauer \& Dumais 1997) is statistically
indistinguishable from that of a
population of non-native English speakers on the same questions.
PMI-IR~(Turney 2001) performed better, but the difference is not
statistically significant.  Jarmasz \& Szpakowicz (2003) give results for a
number of relatively sophisticated thesaurus-based methods that looked
at path length between words in the heading classifications of Roget's
Thesaurus.  Their best scoring method was a statistically significant
improvement over the LSA results, but not over those of PMI-IR.
Terra \& Clarke (2003) studied a variety of corpus-based similarity
metrics and measures of context and achieved a statistical tie with
PMI-IR and the results from Roget's Thesaurus.

To compare directly to these results, we removed the 80 TOEFL
instances from our collection and used the other 351 instances for
training the product rule.  The resulting accuracy
was statistically significantly better than all previously published
results, even though the individual modules performed nearly
identically to their published counterparts.  

\section{Analogies}
\label{turney-s:analogies}
\index{analogies}

Synonym questions are unique because of the existence of
thesauri --- reference books designed precisely to answer queries of
this form.  The relationships exemplified in analogy questions are
quite a bit more varied and are not systematically compiled.  For
example, the analogy question {\small\sf cat}:{\small\sf meow}:: (a)
{\small\sf mouse}:{\small\sf scamper}, (b) {\small\sf bird}:{\small\sf peck}, (c)
{\small\sf dog}:{\small\sf bark}, (d) {\small\sf horse}:{\small\sf groom}, (e)
{\small\sf lion}:{\small\sf scratch} requires that the reader recognize that (c)
is the answer because both (c) and the stem are examples of the
relation ``$X$ is the name of the sound made by $Y$''.  
This type of common sense knowledge is rarely explicitly documented.

In addition to the computational challenge they present, analogical
reasoning is recognized as an important component in cognition,
including language comprehension~(Lakoff \& Johnson 1980) and high level
perception~(Chalmers et al. 1992). 

To study module merging for analogy problems, we collected 374
5-choice instances. We randomly split the
collection into 274 training instances and 100 testing instances.\index{sat@\textsc{sat}}
We next describe the novel modules we developed for attacking analogy
problems and present their results.

\subsection{Modules}

\textbf{Phrase vectors.} 
We wish to score candidate analogies of the
form {\small\sf A}:{\small\sf B}::{\small\sf C}:{\small\sf D} ({\small\sf A} is to {\small\sf B} as
{\small\sf C} is to {\small\sf D}).  The quality of a candidate analogy depends
on the similarity of the relation $R_1$ between {\small\sf A} and {\small\sf B}
to the relation $R_2$ between {\small\sf C} and {\small\sf D}.  The relations
$R_1$ and $R_2$ are not given to us; the task is to infer these
relations automatically.  
Our approach to this task is
to create vectors $r_1$ and $r_2$ that represent features of $R_1$ and
$R_2$, and then measure the similarity of $R_1$ and $R_2$ by the
cosine of the angle between the vectors $r_1$ and $r_2$ (Turney \& Littman 2003a).
We create a vector, $r$, to characterize the relationship between two
words, {\small\sf X} and {\small\sf Y}, by counting the frequencies of 128
different short phrases containing {\small\sf X} and {\small\sf Y}. 
Phrases include ``{\small\sf X} for {\small\sf Y}'', 
``{\small\sf Y} with {\small\sf X}'', ``{\small\sf X} in the {\small\sf Y}'', and ``{\small\sf Y}
on {\small\sf X}''.
We use these phrases as queries to AltaVista and record
the number of hits (matching web pages) for each query.
This process yields a vector of 128 numbers for a pair of
words {\small\sf X} and {\small\sf Y}.  
The resulting vector $r$ is a kind of
\emph{signature} of the relationship between {\small\sf X} and {\small\sf Y}.

\textbf{Thesaurus paths.} 
Another way to
characterize the semantic relationship, $R$, between two words,
{\small\sf X} and {\small\sf Y}, is to find a path through a thesaurus or
dictionary that connects {\small\sf X} to {\small\sf Y} or {\small\sf Y} to {\small\sf X}.
In our experiments, we used the WordNet
thesaurus~(Fellbaum 1998).  We view WordNet as a directed graph and
the Thesaurus Paths module
performed a breadth-first search for paths from {\small\sf X} to {\small\sf Y} or
{\small\sf Y} to {\small\sf X}.  For a given
pair of words, {\small\sf X} and {\small\sf Y}, the module considers all shortest paths
in either direction up to three links.
It scores the candidate analogy by the maximum degree of similarity
between any path for {\small\sf A} and {\small\sf B} and any path for {\small\sf C}
and {\small\sf D}.  The degree of similarity between paths is measured
by their number of shared features.

\textbf{Lexical relation modules.} 
We implemented a set of more
specific modules using the WordNet thesaurus.  Each module checks if
the stem words match a particular relationship in the database.  If
they do not, the module returns the uniform distribution.  Otherwise,
it checks each choice pair and eliminates those that do not match.
The relations tested are: {\bf \small Synonym}, {\bf \small Antonym},
{\bf \small Hypernym}, {\bf \small Hyponym}, {\bf \small
Meronym:substance}, {\bf \small Meronym:part}, {\bf \small
Meronym:member}, {\bf \small Holonym:substance}, and also {\bf \small
Holonym:member}.  

\textbf{Similarity.}
Dictionaries are a natural source to use for solving analogies because
definitions can express many possible relationships and are likely to
make the relationships more explicit than they would be in general
text.  We employed two definition similarity modules: {\bf \small
Similarity:dict} uses {\tt \small dictionary.com} and
{\bf \small Similarity:wordsmyth} uses {\tt \small wordsmyth.net} for definitions.
Each module treats a word as a vector formed from the words in its
definition.  Given a potential analogy
{\small\sf A}:{\small\sf B}::{\small\sf C}:{\small\sf D}, the module computes a vector
similarity of the first words ({\small\sf A} and {\small\sf C}) and adds it to
the vector similarity of the second words ({\small\sf B} and {\small\sf D}).

\begin{table}
\begin{small}
\begin{center}
\begin{tabular}{lcc}
Analogy solvers		&  Accuracy  & Mean likelihood \\
 \hline
Phrase Vectors		& 38.2\% & .2285\\
Thesaurus Paths		& 25.0\% & .1977\\
Synonym			& 20.7\% & .1890\\
Antonym			& 24.0\% & .2142\\
Hypernym		& 22.7\% & .1956\\
Hyponym			& 24.9\% & .2030\\
Meronym:substance	& 20.0\% & .2000\\
Meronym:part		& 20.8\% & .2000\\
Meronym:member		& 20.0\% & .2000\\
Holonym:substance	& 20.0\% & .2000\\
Holonym:member		& 20.0\% & .2000\\
Similarity:dict		& 18.0\% & .2000\\
Similarity:wordsmyth	& 29.4\% & .2058\\
 \hline
all: mixture		& 42.0\% & .2370 \\
all: logarithmic	& 43.0\% & .2354 \\
all: product		& 45.0\% & .2512 \\
 \hline
no PV: mixture		& 31.0\% & .2135 \\
no PV: logarithmic	& 30.0\% & .2063 \\
no PV: product		& 37.0\% & .2207  \\
\end{tabular}
\end{center}
\end{small}
\caption{Comparison of results for merging rules on analogy
problems}  
\label{turney-t:agyprob}
\vspace*{-1em} 
\end{table}

\subsection{Results} 

We ran the 13 modules described above on our set of
training and testing analogy instances, with the results appearing in
Table~\ref{turney-t:agyprob} (the product rule was used to set weights for
computing individual module mean likelihoods).  For the most part, individual module accuracy
is near chance level (20\%), although this is misleading because most
of these modules only return answers for a small subset of instances.
Some modules did not answer a single question on the test set.
The most accurate individual module was the search-engine-based Phrase
Vectors (PV) module.  The results of merging all modules was only a
slight improvement over PV alone, so we examined the effect of
retraining without the PV module.  The product rule resulted in a
large improvement (though not statistically significant) over the best
remaining individual module (37.0\% vs.\ 29.4\% for Similarity:wordsmyth).

We once again examined the result of deducting $1/2$ point for each
wrong answer.  The full set of modules scored 31, 33, and 43 using the
mixture, logarithmic, and product rules.  As in the synonym
problems, the logarithmic and product rules assigned
probabilities more precisely.  In this case, the product rule appears
to have a major advantage.

\section{Conclusion}
\label{turney-s:conclusion}

We applied three trained merging rules to a set of multiple-choice
problems and found all were able to produce state-of-the-art
performance on a standardized synonym task by combining four less
accurate modules.  Although all three rules produced comparable
accuracy, the popular mixture rule was consistently weaker than the
logarithmic and product rules at assigning high probabilities to
correct answers.  We provided first results on a challenging verbal
analogy task with a set of novel modules that use both lexical
databases and statistical information.

In nearly all the tests that we ran, the logarithmic rule and our novel
product rule behaved similarly, with a hint of an advantage for the
product rule.  One point in favor of the logarithmic
rule is that it has been better studied so its
theoretical properties are better understood.  It also is able to
``sharpen'' probability distributions, which the product rule cannot
do without removing the upper bound on weights.  On the other hand,
the product rule is simpler, executes much more rapidly, and is 
more robust in the face of modules returning 
zero probabilities.  We feel the strong showing of the product rule 
proves it worthy of further study. \\

\noindent
{\small {\bf Acknowledgements.}
This research was supported in part by a grant from NASA.
We'd like to thank the students and TAs of Princeton's COS302 in 2001 for
their pioneering efforts in solving analogy problems; Douglas Corey
Campbell, Brandon Braunstein, and Paul Simbi, who provided the first
version of our Thesaurus module; Yann LeCun for scanning help, and
Haym Hirsh, Philip Resnik, Rob Schapire, Matthew Stone, and Kagan
Tumer who served as sounding boards on the topic of module merging.}

\vspace*{-1ex} 


   \cleardoublepage
   \runningheads{Index \ of \ Subjects \ and \ Terms}{Index \ of \ Subjects \ and \ Terms}
\end{document}